\documentclass[runningheads]{llncs}
\usepackage{xcolor}
\usepackage{graphicx}
\usepackage{placeins} % for FloatBarrier command
\usepackage[misc]{ifsym} % for envelope icon

\usepackage[hidelinks]{hyperref}

\usepackage[ruled, lined, longend]{algorithm2e}
\usepackage{amssymb,amsmath,array}
\usepackage{mathtools}
\usepackage{upgreek}

\usepackage{setspace}

\SetCommentSty{mycommentfont}

\usepackage{tabularx}
\usepackage{booktabs}
\usepackage{multirow}

\newcolumntype{P}[1]{>{\centering\arraybackslash}p{#1}}
\newcolumntype{R}{>{\raggedleft\arraybackslash}X}
\newcolumntype{C}{>{\centering\arraybackslash}X}

\renewcommand{\vec}[1]{\mathbf{#1}}

\newcommand{\netinput}{\vec{x}}
\newcommand{\netinputpred}{\tilde{\vec{x}}}
\newcommand{\forwardpred}{f_{p}}
\newcommand{\forwardcls}{f_{c}}

\newcommand{\hidpred}{\vec{h}_{p}}
\newcommand{\hidcls}{\vec{h}_{c}}

\newcommand{\outcls}{\vec{y}_{c}}

\begin{document}

\title{Early Recognition of Ball Catching Success in Clinical Trials with RNN-Based Predictive Classification}

\titlerunning{Early Recognition of Clinical Catching Trials}

\author{Jana Lang\inst{1}(\Letter) \and
Martin A. Giese\inst{1} \and
Matthis Synofzik\inst{2} \and
Winfried Ilg\inst{1}\thanks{Shared last authors} \and
Sebastian Otte\inst{3}\protect\footnotemark[1]}

\authorrunning{J. Lang et al.}

\institute{Section for Computational Sensomotorics, Department of Cognitive Neurology, Centre for Integrative Neuroscience \& Hertie Institute for Clinical Brain Research, University Clinic Tübingen, Germany \\
\email{jana.lang@uni-tuebingen.de} \and
Department of Neurodegeneration, Hertie Institute for Clinical Brain Research \& Centre for Neurology, University Clinic Tübingen, Germany \and
Neuro-Cognitive Modeling Group, University of Tübingen, Germany}
\maketitle  % typeset the header of the contribution
\begin{abstract}

Motor disturbances can affect the interaction with dynamic objects, such as catching a ball. A classification of clinical catching trials might give insight into the existence of pathological alterations in the relation of arm and ball movements. Accurate, but also early decisions are required to classify a catching attempt before the catcher's first ball contact. To obtain clinically valuable results, a significant decision confidence of at least 75\,\% is required. Hence, three competing objectives have to be optimized at the same time: accuracy, earliness and decision-making confidence. 
Here we propose a coupled classification and prediction approach for early time series classification: a predictive, generative recurrent neural network (RNN) forecasts the next data points of ball trajectories based on already available observations; a discriminative RNN continuously generates classification guesses based on the available data points and the unrolled sequence predictions. We compare our approach, which we refer to as \emph{predictive sequential classification} (PSC), to state-of-the-art sequence learners, including various RNN and temporal convolutional network (TCN) architectures. On this hard real-world task we can consistently demonstrate the superiority of PSC over all other models in terms of accuracy and confidence with respect to earliness of recognition. Specifically, PSC is able to confidently classify the success of catching trials as early as 123 milliseconds before the first ball contact. We conclude that PSC is a promising approach for early time series classification, when accurate and confident decisions are required.

\end{abstract}

\section{Introduction}

Patients suffering from neurodegenerative or neurodevelopmental disorders, including Spinocerebellar Ataxia and Autism Spectrum Disorder, are often impaired in the interaction with dynamic objects, for instance when catching a ball. Ball catching requires an intact perception-action coupling and the ability to anticipate the trajectory of an oncoming ball \cite{whyatt2012}. It is suggested that dysfunctions of predictive control which result in alterations of preparatory arm movements are the leading cause of catching impairments in these diseases \cite{franklin2011}. In this paper, we aim to recognize changes in the relation of arm and ball movements that are predictive for the success of catching trials before the first ball contact of the catcher. To ensure clinically valuable results, we choose a confidence threshold of 75\% which is commonly used for two-alternative forced choice tasks in psychophysical studies \cite{ulrich2009}.

Hence, the problem at hand can be formulated as an \textit{early time series classification} task with increased confidence requirements. Early time series classification is referred to as making classifications as early as possible, while maintaining a high classification accuracy \cite{santos2016}. It naturally evokes a trade-off between earliness and accuracy of classifications. Different approaches have been applied to the problem of early time series classification, including convolutional neural networks and reinforcement learning \cite{wang2016,martinez2018}. Surprisingly, despite the widespread application of recurrent neural network (RNN) models to sequential problems, they have rarely been used for the early classification of time series. Recent work, however, suggests that gated recurrent units can handle missing values in multivariate time series \cite{che2018}. Moreover, first promising results have been achieved applying long short-term memory (LSTM) models for early classification in agricultural monitoring \cite{russwurm2019}. These approaches leave aside the confidence of decision-making, which is an important factor in various applications.

Recently, it has been shown that predictive RNNs can be employed to efficiently generate goal-directed, anticipatory behavior to support decision-making \cite{otte2017}. Therefore, we present a novel RNN-based approach that simultaneously optimizes accuracy, confidence and earliness in time series classification. Our approach, which we refer to as \textit{predictive sequential classification} (PSC), incorporates two different specialized RNN models into one coupled arrangement. The first model, a predictive, generative RNN, forecasts the next data points of a time series based on already available observations. The second model, a discriminative RNN, continuously generates classification guesses based on the available data points and the unrolled sequence predictions of the first model. We compare our approach to several state-of-the-art sequence learners, including various RNN and temporal convolutional network (TCN) architectures using a motion dataset containing two-dimensional trajectories of healthy and pathological ball catching attempts. At test time, all models are confronted with incomplete catching trials of different lengths. We evaluate all architectures with regard to the accuracy of the final decision, the level of decision confidence, and the earliness of decision-making.

\vfill

\section{Predictive Sequential Classification}

\textit{Time series classification} describes the task of assigning one of two (binary classification) or one of multiple (multi-class classification) labels to a time series $S$, where $S$ is defined as an ordered, uniformly spaced temporal sequence of $T$ vectors \cite{esling2012}:

\begin{equation}
\label{eqn:timeseries}
%S = (s^0, s^1, … , s^{T-1})
S = (\netinput^{1}, \netinput^{2}, \ldots, \netinput^{T})
\end{equation}

The here considered time series are \textit{multivariate}, i.e. they contain more than one feature for each time step. When processing time series, two different types of processing modes can generally be distinguished. The first mode, called \textit{many-to-one} (MTO) processing, takes an input sequence and outputs a single class label after consuming the entire input sequence. \textit{Many-to-many} (MTM) processing, on the other hand, produces a class label for multiple steps (typically for every time step) of the input sequence, i.e. both input and output are sequences. In contrast to other types of neural networks, sequence learners, such as the recurrent neural network (RNN), expect the data to be temporally highly correlated and in sequential order \cite{goodfellow2016}. By introducing circular connections (\textit{recurrences}), RNNs allow past inputs to influence future time steps. However, in practical applications, vanilla RNNs are largely replaced by long short-term memory (LSTM) networks. This extension of RNNs overcomes the vanishing gradient problem and makes the learning of long-term dependencies possible \cite{hochreiter1997}.

\textit{Early time series classification} is targeted at making accurate classifications based on incomplete, instead of full-length time series. To compensate for the missing time interval we propose a novel RNN-based approach to early time series classification that equips a sequence classifier with predictive power. The \textit{predictive sequential classification} (PSC) approach entails both a predictive, generative LSTM that forecasts the next data points of a time series based on already available observations, as well as a discriminative LSTM, which continuously generates classification guesses based on the available data points and the unrolled sequence predictions. Both models are trained separately on their respective tasks. At test time the models come together to make a predictive classification guess.

\begin{figure}[t!]
    \centering
    \includegraphics{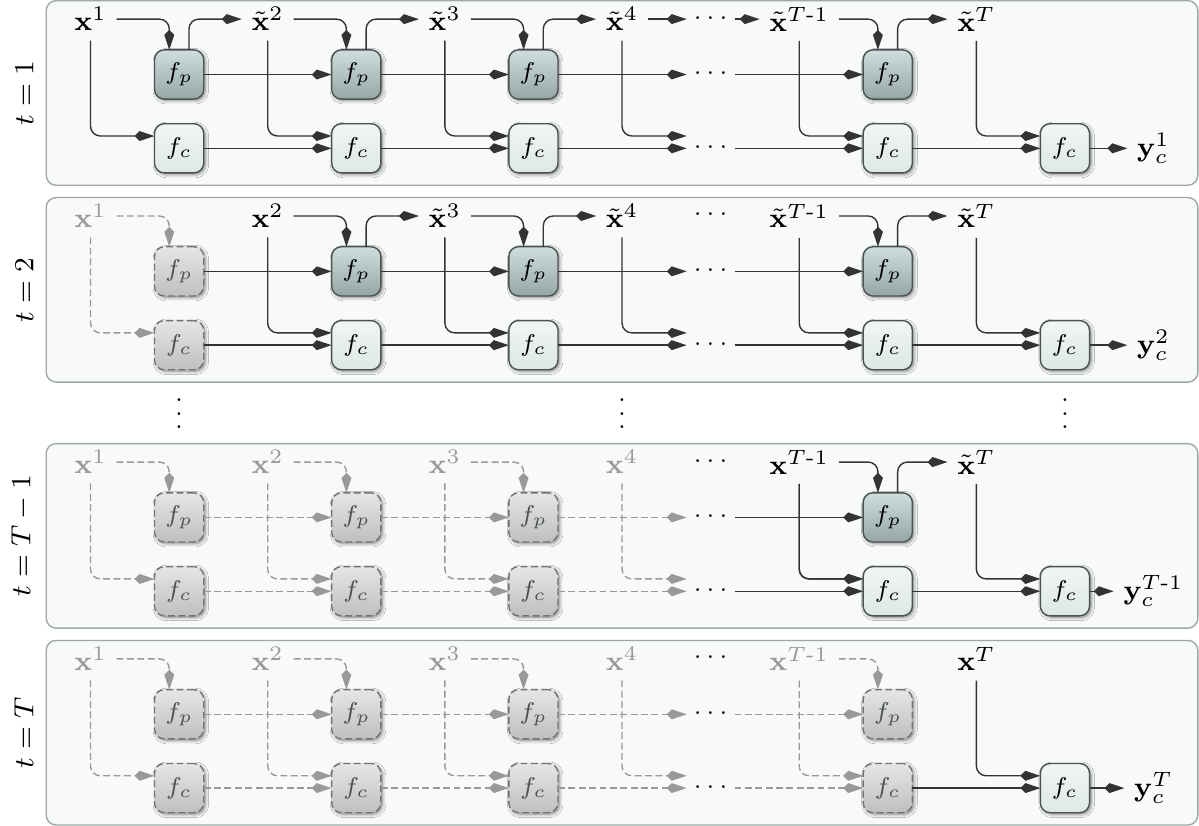}
    \caption{Predictive sequential classification. A predictive RNN forecasts the remaining trajectories based on past observations. The classifier combines both to make informed classification guesses. Grayed out boxes indicate previously computed states.
    }
    \label{fig:psc}
\end{figure}

With every incoming observation $\netinput^{t}$, the amount of available observations increases. Based on these data points, the predictor sequentially forecasts the remaining $T-t$ data points of the time series. Each predicted observation $\netinputpred^{t}$ is used to make a predictive classification. Finally, the classification output $\outcls^{t}$ is updated with the last predictive classification $\outcls^{T}$ (\autoref{alg:predcls}). For every history size, the classifier aggregates already available observations and predicted observations to make a predictive classification guess. When all data points of the time series are available, PSC defaults to a vanilla sequence classifier (\autoref{fig:psc}). In the following experiments, we show that PSC is superior to state-of-the-art sequence classifiers for the task of early and confident time series classification. In an additional study, we investigate the importance of the two-model design for PSC, revealing that directly including a predictive objective into a single model even harms the classification performance.

\begin{algorithm}[t!]
    %\setstretch{1.0}
    
    \caption{Predictive Sequential Classification}
    \label{alg:predcls}
    
    \renewcommand{\BlankLine}{\vskip 5pt}
    \DontPrintSemicolon
    \SetNoFillComment
    
    \tcc{Initialize hidden states}
    $\hidcls^{0}$ $\leftarrow$ $\vec{0}$, $\hidpred^{0}$ $\leftarrow$ $\vec{0}$
    %\BlankLine
    \tcc{Loop over incoming observations}
    \For(){$t$ $\leftarrow$ $1$ \KwTo $T$}{
        \tcc{Update classifier and predictor with current input}
        $\outcls^{t}, \hidcls^{t}$ $\leftarrow$ $\forwardcls(\netinput^{t},\hidcls^{t-1})$\\
        $\netinputpred^{t+1}, \hidpred^{t}$ $\leftarrow$ $\forwardpred(\netinput^{t}, 
        \hidpred^{t-1})$\\
        %\BlankLine
%        y_g' = y_g
%        h_c' = h_c
%        h_g' = h_g
%        y_g' = y_g
        
        %\BlankLine
       
        \tcc{Unroll sequence prediction  and predictive classification}
        \For(){$t'$ $\leftarrow$ $t + 1$ \KwTo $T$}{
            $\outcls^{t'}, \hidcls^{t'}$ $\leftarrow$ $\forwardcls(\netinputpred^{t'},\hidcls^{t' - 1})$\\
            $\netinputpred^{t'+1}, \hidpred^{t'}$ $\leftarrow$ $\forwardpred(\netinputpred^{t'}, 
            \hidpred^{t'-1})$\\
        }

        %\BlankLine
        \tcc{Use predictive classification as current classifier output}
        $\outcls^{t}$ $\leftarrow$ $\outcls^{T}$
    }
    \BlankLine
    {\scriptsize
    \setstretch{1.05}
    Variables:
    
    \begin{tabular}{llllll}
    $t$ & : & current time step                   & $\hidcls^t$ & : & the classifier's hidden state\\
    $t'$ & : & time step within prediction loop~~ & $\hidpred^t$ & : & the predictor's hidden state\\
    $T$ & : & sequence length                       & $\forwardcls$ & : & the classifier's forward pass function\\
    $\netinput^{t}$ & : & observation at time $t$ & $\forwardpred$ & : & the predictor's forward pass function\\
    $\netinputpred^{t}$ & : & predicted observation for time $t$ & $\outcls^{t}$ & : & classification output for time $t$\\
    \end{tabular}
    }
\end{algorithm}

\FloatBarrier
\section{Experimental Setup}
\subsection{Data}
The data sample comprised 63 videos of one-handed ball catching attempts by 11 healthy subjects, 13 children with Autism Spectrum Disorder and 10 patients with Spinocerebellar Ataxia. All experiments were admitted by the ethical committee of the University Clinic of Tübingen. Of the 1975 recorded catching trials, 1082 attempts were successful, and 893 attempts were unsuccessful. 965 trials were caught with the right hand and 1010 trials with the left hand. Videos were recorded at a frame rate of 100 frames per second. The two-dimensional position of the catcher’s arm and hand joints and the trajectory of the ball were captured with two deep learning frameworks for pose estimation \cite{mathis2018,cao2019}. We extracted 18 hand and arm features describing the motion of the shoulder, the elbow, the wrist, the ball of the hand, each fingertip on the relevant body side, and two features specifying the ball center for each video frame.

\subsection{Preprocessing}

A Savitzky-Golay filter was applied to the arm and hand marker trajectories between the start and the end of each trial to smooth flickering noise. All catching trials were provided with a binary label specifying the success of the attempt (1: catch, 0: drop). An attempt was only counted as successful if the ball was caught at the first try. Two types of dropping behavior were observed, where the ball either jumped off the catcher's hand or the catcher completely missed the ball. Learning absolute coordinates can lead to overfitting, since a slightly shifted starting position already leads to different absolute coordinates, while the relative difference can remain the same. Therefore, absolute coordinates were converted to relative coordinates by taking the difference between coordinates of two subsequent frames. Since every catching attempt varied in length, we truncated all sequences to the shortest sequence length, taking into account the trade-off between comparability of models, accuracy and sufficiently long sequence lengths. Hence, the first $f$ frames of all sequences were removed to obtain the length of the shortest sequence (60 frames, i.e. 600 milliseconds (ms)), with $f$ as the difference between the individual sequence lengths and the shortest sequence length. The full dataset was then randomly split into subsets for training (60\,\%), validation (20\,\%) and testing (20\,\%). All subsets were normalized using mean and standard deviation of the training subset.

\subsection{Models}
We compare PSC to other LSTM models and temporal convolutional networks (TCN) \cite{hochreiter1997,bai2018}. All models are trained using the Adam optimizer with standard parameters ($\eta = 0.001$ (learning rate), $\beta_1 = 0.9$ and $\beta_2 = 0.999$, $\epsilon = 10^{-7}$) and the binary cross-entropy (BCE) loss.

\textbf{LSTM Models}:
All LSTM models contain two LSTM layers with 64 hidden units, one dropout layer after each LSTM layer, recurrent dropout of 20\% and a fully-connected output layer with sigmoid activation. They are trained on full batches. The weights of LSTM layers are initialized according to Xavier uniform initialization. Recurrent weights are initialized in an orthogonal manner. The many-to-one model (MTO-LSTM) maps the input sequence to a single binary classification. A dropout of 50\% is applied. It is trained for 250 epochs. The many-to-many model (MTM-LSTM) produces classification guesses for each time step of the input sequence. A dropout of 40\% is applied. MTO-LSTM and MTM-LSTM have 54,849 trainable parameters each. The hybrid model (HYB-LSTM) extends the fully-connected layer of MTM-LSTM by an additional branch for trajectory prediction with linear activation. At each time step $t$, HYB-LSTM simultaneously produces a classification guess for step $t$ and a prediction for arm, hand and ball trajectories at $t+1$. It is trained on an equally-weighted additive loss, consisting of the BCE for classification outputs and the mean squared error for regression outputs. HYB-LSTM includes 56,149 parameters. Finally, PSC-LSTM realizes our \textit{predictive sequential classification} approach. For classification, we use the above described MTM-LSTM model. For prediction, we train a separate LSTM on trajectory prediction which resembles the prediction branch of HYB-LSTM. MTM-LSTM, HYB-LSTM and the ancillary prediction network of PSC-LSTM are trained for 200 epochs.

\textbf{TCN Models}:
Three TCN models are implemented, each of which covers a different receptive field size (10, 30 or 60 steps). The size of the receptive field determines the number of residual stacks, the size of the kernels and the dilation factors used (\textit{receptive field size = number of stacks * kernel size * last dilation factor}). We examined different combinations of these factors for each receptive field size and selected the architecture that yielded the highest validation accuracy. TCN-10 is composed of one residual block, 32 filters of size 2, dilations of 1 and 5 and a dropout rate of 0.2. It is trained on batches of 32 samples and has 8,257 trainable parameters. TCN-30 holds 3 residual blocks, 20 filters of size 2, dilations of 1 and 5 and a dropout rate of 0.3 and is trained on batches of 64 samples. It contains 9,861 trainable parameters. Finally, TCN-60 contains 2 residual blocks, 20 filters of size 2, dilations of 1, 5, 10 and 15 and a dropout rate of 0.3 and is trained on batches of 64 samples. It comprises 13,141 trainable parameters. All TCN models are trained for 500 epochs and contain one final fully-connected layer with sigmoid activation to produce class probabilities for each time step. A He normal initializer is used for TCN kernels.

\subsection{Evaluation Metrics}
All models were tested on an unseen subset of the original data containing 395 randomly selected catching attempts. The models are rated according to the degree of correctness of a classification, as well as the earliness of decision-making.

\textbf{Accuracy}:
 We evaluate the models based on the percentage of correctly classified test trials (the percent accuracy) given different sizes of past history. We apply two different thresholds to measure the confidence of a prediction. The 50\,\% confidence threshold rounds the final model output to 1 or 0, symbolizing a catch or a drop, respectively and compares it to the target label. When applying a 75\,\% confidence threshold, however, trials are only considered if the final model output either exceeds 0.75 or undershoots 0.25. In the first case, the binary output is set to 1, in the latter case to 0. Trials with model outputs between 0.25 and 0.75 are disregarded. The binary output is again compared to the target label.

\textbf{Ball-Hand Distance}:
The distance between the catcher's hand and the ball can be a first indicator to determine whether the ball will be caught or not. The metric is used to qualitatively evaluate the performance of a model by visually comparing trends in the prediction curve with the ball-hand distance over time. It is defined as the Euclidian distance between the absolute two-dimensional positions of the marker at the base of the relevant hand and at the center of the ball at a given time step $t$. The ball-hand distance increases when ball and hand move farther apart, while it decreases when the markers are moving closer together. However, since the ball-hand distance only considers the base of the hand, it does not necessarily give insight into the ball being grasped or not.

\textbf{Mean Time to (Correct) Decision}:
We introduce a novel metric to quantitatively evaluate the earliness of decision-making. The time to decision ($TTD_i$) for a given input sequence (sample) $i$ is defined as the time step when the model makes a final decision without switching decisions afterwards. A final decision is defined as a model prediction of larger than 0.75 or smaller than 0.25, regardless of the correctness. Values between 0.25 and 0.75 are treated as indecisive and excluded in further calculations. 

$\forall \vec{y}_{i} : y_{i}^{T} \geq \theta_{hi} \vee  y_{i}^{T} \leq \theta_{lo}$

\begin{equation}
\label{eqn:ttd}
\begin{aligned}
	TTD(Y) = \max \bigl\lbrace 1 \leq t \leq T \mid 
	& \left(y^{t-1} < \theta_{hi} \wedge  y^{t} \geq \theta_{hi}\right)
    \\
     \vee &
	\left(y^{t-1} > \theta_{lo} \wedge  y^{t} \leq \theta_{lo}\right)
	\bigr\rbrace
\end{aligned}
\end{equation}

The time to correct decision ($TTcD_i$) for a given sample $i$ equals the time to decision if the final decision is correct, i.e. if the binary model output $y_{bin_i}$ after applying a 75\,\% confidence threshold equals the target label $z_i$. Otherwise, the $TTcD_i$ is undefined:

\begin{equation}
\label{eqn:ttcd}
	TTcD(Y) =
	\begin{cases}
		\phantom{unde}TTD(Y),\qquad	 y_{bin_i} = z_i\\
		undefined,\qquad y_{bin_i} \neq z_i\\
	\end{cases}
\end{equation}

The mean time to decision ($MTTD$) and the mean time to correct decision ($MTTcD$) are defined as the arithmetic mean of all defined $TTD_i$ and $TTcD_i$, respectively, for $i$ in the number of data samples $N$. A low MTTcD indicates that a model can make the correct decision early in the time sequence, i.e. based on a small size of available past history, whereas a low MTTD only implies that a model tends to make decisions early, but not necessarily correctly.

\section{Results and Discussion}
\begin{figure}[b!]
\includegraphics[width=\textwidth]{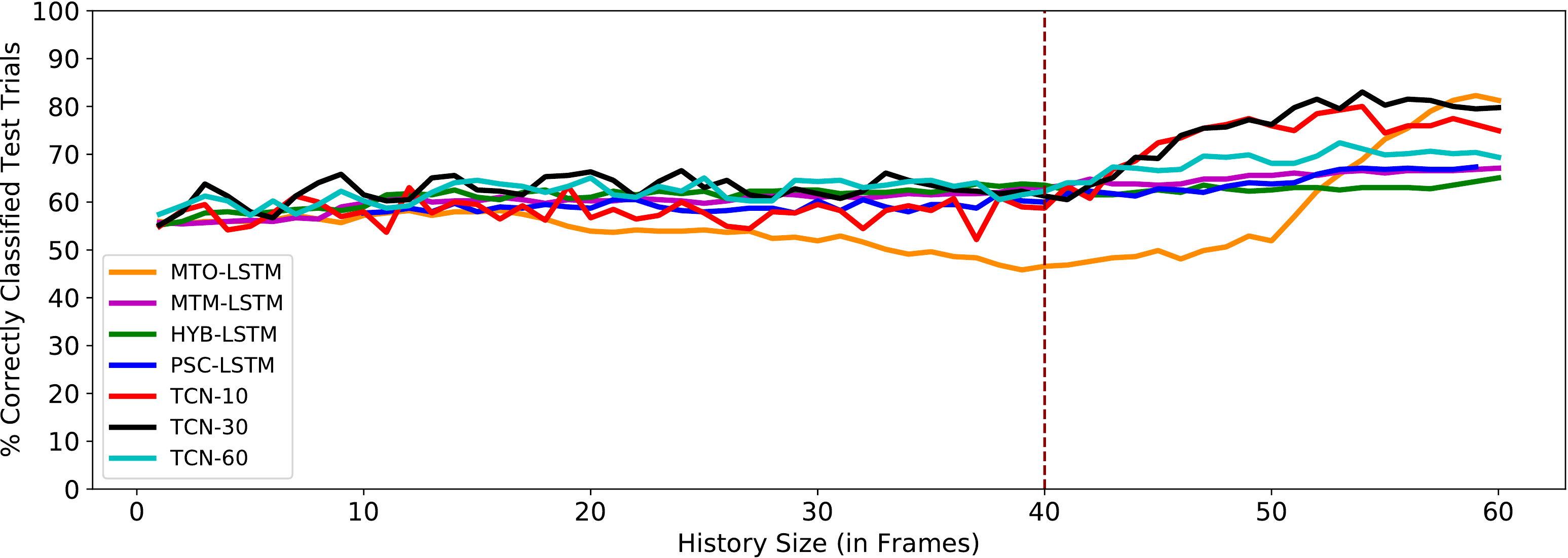}
\caption{Model accuracy applying a 50\,\% decision confidence threshold. The figure illustrates the percentage of correctly classified test trials for increasing sizes of past history for all models. The vertical red line denotes the point of the catcher's first ball contact.} \label{fig:50accuracy}
\end{figure}

All models are confronted with incomplete trajectories of catching trials. Therefore, the ready-trained models are used to make a classification based on increasing sizes of known past history. The presented results are average values based on ten repetitions with random data splits. To evaluate the correctness of the final classification, two different confidence thresholds are applied. \autoref{fig:50accuracy} depicts the percentage of correctly classified test samples for all models assuming a decision confidence threshold of 50\,\%. This figure shows that all models start with an accuracy above chance level. MTO-LSTM is the only model that falls below chance level for history sizes between 34 and 47 time steps. However, it reaches the highest accuracy of 81.27\,\% when the entire history is known, i.e. after 60 time steps of accumulated history. PSC-LSTM only starts to make predictions after a warm-up phase of ten time steps where history is accumulated. When the entire past history is available, it defaults to MTM-LSTM. Hence, both models reach the same final accuracy of 67.08\,\% after 60 time steps. However, MTM-LSTM performs slightly better over time applying a 50\,\% confidence threshold. HYB-LSTM demonstrates constant accuracies around 60\,\% with an increase to 65.06\,\% for the complete sequence. When applying a 50\,\% confidence threshold, the TCN models outperform the LSTM models, especially with larger history sizes, without dropping below chance.

\begin{figure}[t!]
\includegraphics[width=\textwidth]{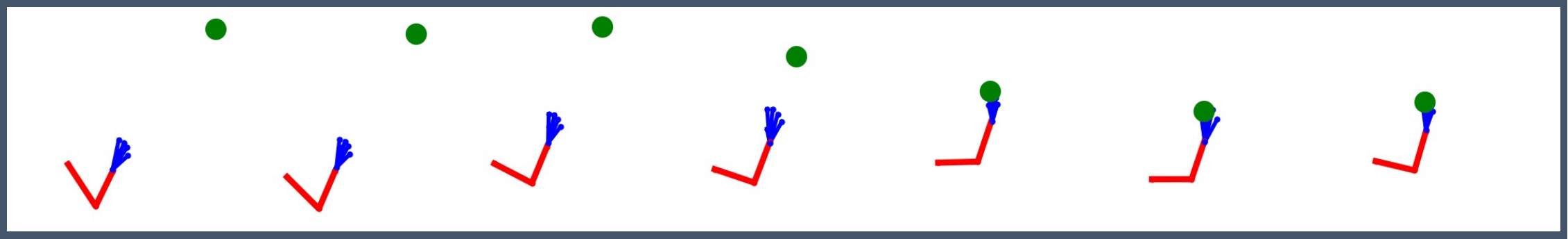}
\caption{Animated two-dimensional trajectories of the catcher's arm (red) and hand (blue) and the ball (green) for a successful sample trial over time. The figure illustrates the trajectories every ten time steps, starting with step 0. The fifth pose denotes the point of the catcher's first ball contact.} \label{fig:animation}
\end{figure}

\begin{figure}[t!]
\includegraphics[width=\textwidth]{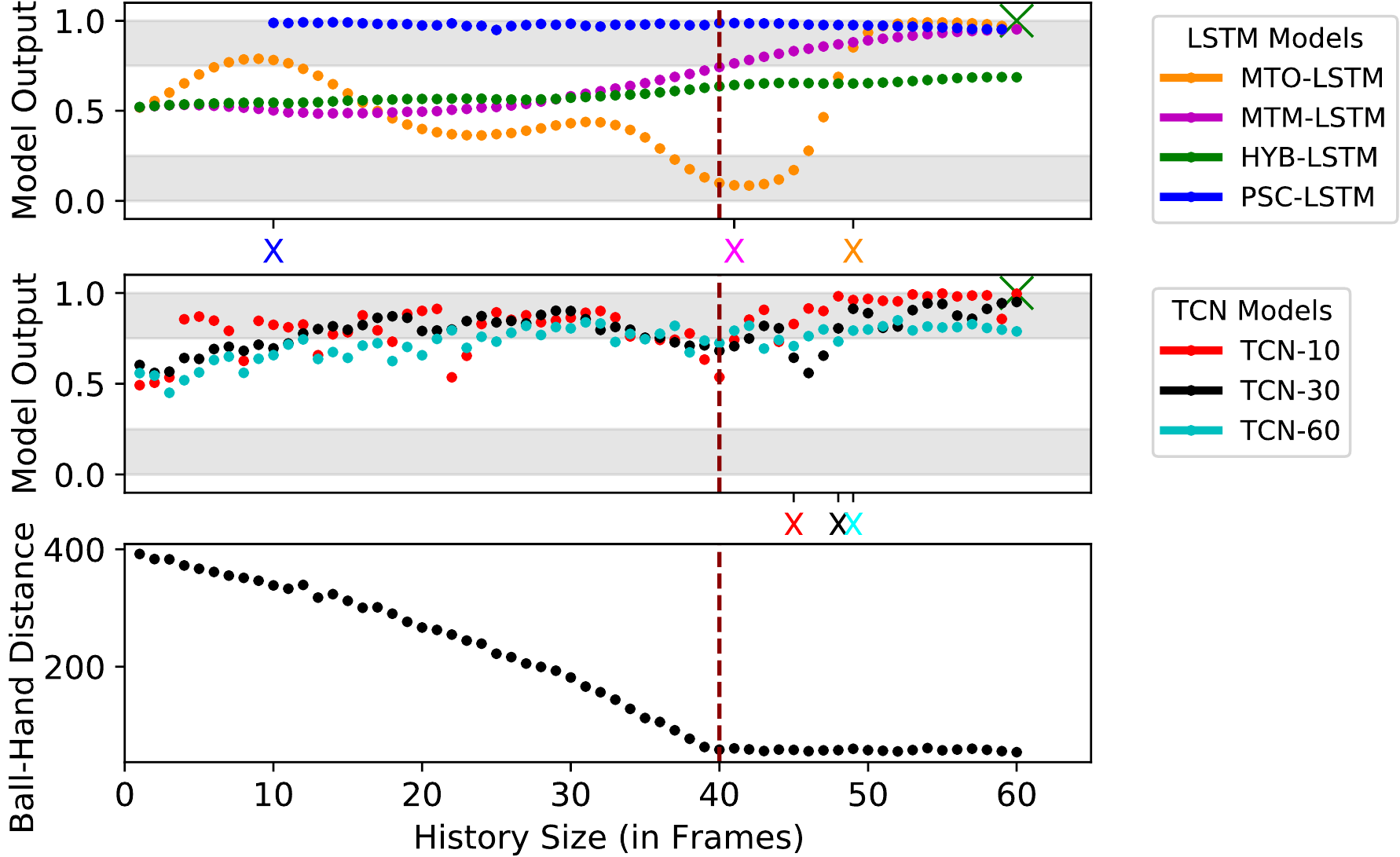}
\caption{Predictions of all models for a successful sample trial (cf. Figure \ref{fig:animation}). The uppermost sub-figure illustrates the model classifications of all LSTM models for increasing history sizes. The sub-figure in the middle shows the predictions of TCN models. In both figures, the green cross denotes the correct label for the selected sample trial. Colored crosses below the sub-figures mark the TTcD for the model of the corresponding color. The vertical red line at step 40 shows the point of the catcher's first ball contact. The bottom figure shows the ball-hand distance over time for the selected trial.}\label{fig:trial}
\end{figure}

\begin{figure}[t!]
\includegraphics[width=\textwidth]{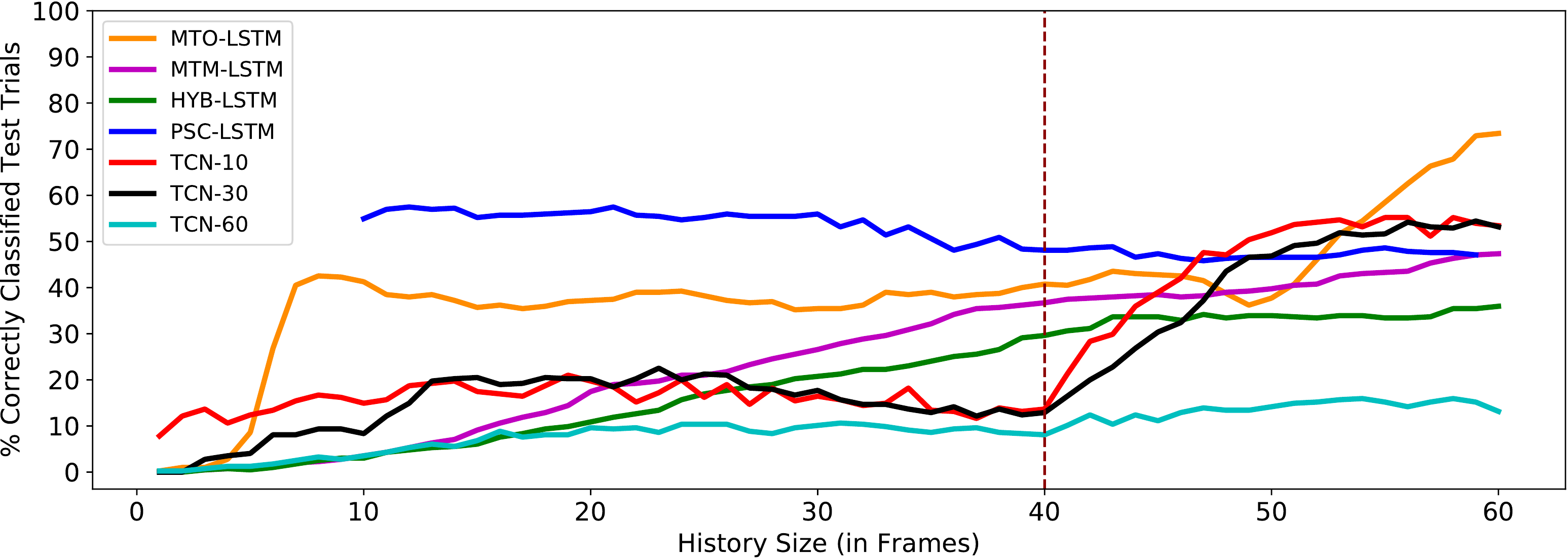}
\caption{Model accuracy applying a 75\,\% decision confidence threshold. The figure illustrates the percentage of correctly classified test trials for increasing sizes of past history for all models. The vertical red line denotes the point of the catcher's first ball contact.} \label{fig:75accuracy}
\end{figure}

\begin{table}[t!]
\caption{Comparison of the mean times to (correct) decision. Numbers in brackets denote the temporal distance to the catcher's first ball contact at frame 40 (after 400 ms). Negative distances represent classifications before the initial ball contact, and positive distances indicate classifications after the catcher's first ball contact.}
\label{tab:mttd}
\scriptsize
\begin{tabularx}{\linewidth}{CCCCC}
\toprule
\multirow{2}{*}{\textbf{Model}} & \textbf{No.} & \textbf{MTTD} &
\textbf{No. correct} & \textbf{MTTcD} \\ 
& \textbf{decisions} & [ms]
& \textbf{decisions} & [ms] \\ 
\midrule
MTO-LSTM                           & 327                       & 475.5 (+75.5)         & 290                               & 475.9 (+75.9)         \\
MTM-LSTM                           & 258                       & 316.1 (-83.9)         & 187                               & 313.3 (-86.7)          \\
HYB-LSTM                           & 191                       & 348.4 (-51.6)         & 142                               & 334.4 (-65.6)          \\
\textbf{PSC-LSTM}                            & 257                       & 277.7 \textbf{(-122.3)}         & 186                               & 276.9 \textbf{(-123.1)}          \\
TCN-10                             & 242                       & 524.5 (+124.5)         & 211                               & 516.2 (+116.2)          \\
TCN-30                             & 234                       & 482.1 (+82.1)         & 210                               & 476.0 (+76.0)          \\
TCN-60                             & 60                        & 495.3 (+95.3)         & 52                                & 481.3 (+81.3)         \\
\bottomrule
\end{tabularx}%
\end{table}

However, when considering a 75\,\% confidence threshold, the picture changes significantly. \autoref{fig:animation} depicts selected steps of a sample trial where the subject was able to catch the ball. In \autoref{fig:trial} the corresponding model predictions and the ball-hand distance for this trial are displayed. The latter shows a clear decrease until step 40, the point of time when the ball is first touched. Since the ball is successfully caught, the ball-hand distance stays minimal after the catcher's first ball contact. The two upper sub-figures depict the process of decision-making along the course of the trial. The point at which a model reaches the gray-shaded decision area between 0.75 and 1.0 without decision-switching afterwards, is denoted as the TTcD of the corresponding model. PSC-LSTM already commits to the final correct decision at the first prediction attempt after ten steps. MTM-LSTM expresses a continuously increasing confidence with a TTcD of 41, while HYB-LSTM does not reach a final confident decision and stays below 0.75. MTO-LSTM exhibits decision-switching and only comes to a final confident decision at step 49. Compared to the LSTM models, all TCN models demonstrate a more noisy prediction curve and high TTcDs (TCN-10: 45, TCN-30: 48, TCN-60: 49). PSC-LSTM is the only model which can predict the success of the trial before the ball-hand distance reaches 0.

Figure \ref{fig:75accuracy} illustrates the resulting percentage of correctly classified trials across all test samples for different history sizes. First, it can be observed that PSC-LSTM is the only model which is capable of correctly classifying more than half of the test samples with history sizes smaller than 47 steps. For all other models, there is a vast discrepancy between \autoref{fig:50accuracy} and \autoref{fig:75accuracy} which implies that most model classifications fall between 0.25 and 0.75, especially for smaller history sizes. When comparing the LSTM models, MTO-LSTM model again performs best when the entire history is available. This indicates that MTO-LSTM does not learn to make decisions early, but rather waits for the final frames. For smaller history sizes before the first ball contact, PSC-LSTM is the dominant model. Ultimately, the accuracy slightly drops, since it converges to the accuracy of MTM-LSTM. Furthermore, there is an imbalance in the data set, containing a large percentage of jump-off trials where the success of a trial can only be assessed in the last frames, potentially hindering the prediction. HYB-LSTM which incorporates classification and prediction capabilities in one model is outpaced by MTM-LSTM. TCN-60 sticks with uncertain decisions between 0.25 and 0.75 for most of the trials, while TCN-10 and TCN-30 start to make more confident decisions at the moment of the catcher's first ball contact after 40 time steps (400 ms). This increase in accuracy can potentially be attributed to overfitting to the final frames, which is further supported by diverging training and test accuracies.

This observation is confirmed by the low percentage (15\,\%) of confidently classified catching trials made by TCN-60 (\autoref{tab:mttd}). The highest percentage of correct decisions is made by TCN-30 and MTO-LSTM. However, both models rely on late information, while PSC-LSTM achieves the highest accuracy before the first ball contact. Considering that the first ball contact occurs at time step 40, neither the TCN models (MTTcD for TCN-10: 516.2, TCN-30: 476.0, TCN-60: 481.3), nor MTO-LSTM (MTTcD: 475.9) are capable of correctly and confidently classifying catching trials before the outcome is visually observable. The prevailing models are the recurrent networks trained on many-to-many classification. The best performance by far is achieved by PSC-LSTM which can confidently classify trials already 123 ms before the first ball contact.

HYB-LSTM performs slightly worse than MTM-LSTM, but can still make a classification 65.6 ms before the ball is first touched. In contrast to Hüsken and Stagge who argue that incorporating an additional prediction task into a classification RNN improves the learning process, we show that it does not have a positive effect on the earliness of decision-making \cite{husken2003}. However, the outsourced prediction approach of PSC-LSTM is superior to both the embedded prediction approach of HYB-LSTM, as well as the pure classification approach of MTM-LSTM. Hence, the inclusion of prediction capabilities into a classification model seems to harm classification performance. This finding can be a first indicator for the existence of different, possibly antagonistic internal representations and learning strategies of RNNs trained on classification versus regression tasks.

\FloatBarrier

\section{Conclusion}
In this paper we introduced a novel RNN-based approach for early and confident time series classification: the \textit{predictive sequential classification} (PSC). We evaluated our approach in comparison to state-of-the-art sequence learners on the early recognition of clinical ball catching trials. We consistently demonstrate the superiority of PSC over all other LSTM and TCN models in terms of the earliness and accuracy of decisions under a high confidence threshold. Specifically, PSC can on average make a final decision as early as 123 ms before the catcher's first ball contact. Hence, we show that ancillary prediction models clearly benefit classification performance. However, incorporated prediction capabilities seem to interfere with classification skills and ultimately hurt classification performance. Our findings show that PSC with its two-model design can simultaneously optimize accuracy, earliness and confidence of decision-making, thus constituting a promising approach for early and confident time series classification in manifold applications.

\subsubsection{Acknowledgments}
The authors thank Tobias Renner and Gottfried Barth (Department of Child and Adolescent Psychiatry, Psychosomatics and Psychotherapy, University Hospital of Psychiatry and Psychotherapy, Tübingen, Germany) for their support on the assessment of catching movements in children with Autism Spectrum Disorders. The authors thank the International Max Planck Research School for Intelligent Systems (IMPRS-IS) for supporting JL. Additional support was provided by BMG, project SStepKiZ, and by ERC 2019-SyG-RELEVANCE-856495 to MG.

\bibliographystyle{splncs04}
\bibliography{catching}

\end{document}